\documentclass[10pt, conference, letterpaper]{IEEEtran}
\usepackage[font=footnotesize]{caption}
\IEEEoverridecommandlockouts
\usepackage{cite}
\usepackage{amsthm}
\usepackage{amsmath,amssymb,amsfonts}
\usepackage[cmintegrals]{newtxmath}
\usepackage{enumerate}
\usepackage{caption}
\usepackage{bm}
\usepackage{multicol} 
\usepackage{multirow}
\usepackage{booktabs}
\usepackage{graphicx}
\usepackage{array}
\usepackage{textcomp}
\usepackage{xcolor}
\usepackage{url}
\usepackage{tabularx}
\usepackage{color}

\usepackage{xcolor}
\usepackage{ulem}

\usepackage{hyperref}
\usepackage{diagbox}
\usepackage{subfigure} 
\usepackage{amsmath}
\usepackage{amsfonts}
\usepackage{algorithm}
\usepackage{algorithmic}
\usepackage{tikz}
\usetikzlibrary{positioning, arrows.meta} 

\makeatletter

\newtheorem{Assumption}{Assumption}

\begin{document}
\title{Distributed Dynamic Invariant Causal Prediction in Environmental Time Series}  
 \author{Ziruo Hao,
Tao~Yang,~\IEEEmembership{Member,~IEEE,}
Xiaofeng~Wu,
and~Bo~Hu,~\IEEEmembership{Member,~IEEE}      

}

\maketitle
\begin{abstract}
The extraction of invariant causal relationships from time series data with environmental attributes is critical for robust decision‐making in domains such as climate science and environmental monitoring. However, existing methods either emphasize dynamic causal analysis without leveraging environmental contexts or focus on static invariant causal inference, leaving a gap in distributed temporal settings. In this paper, we propose Distributed Dynamic Invariant Causal Prediction in Time-series (DisDy-ICPT), a novel framework that learns dynamic causal relationships over time while mitigating spatial confounding variables without requiring data communication. We theoretically prove that DisDy-ICPT recovers stable causal predictors within a bounded number of communication rounds under standard sampling assumptions. Empirical evaluations on synthetic benchmarks and environment-segmented real-world datasets show that DisDy-ICPT achieves superior predictive stability and accuracy compared to baseline methods A and B. Our approach offers promising applications in carbon monitoring and weather forecasting. Future work will extend DisDy-ICP to online learning scenarios.
\end{abstract}


\begin{figure*}[!t]
\centering
\includegraphics[width=7in]{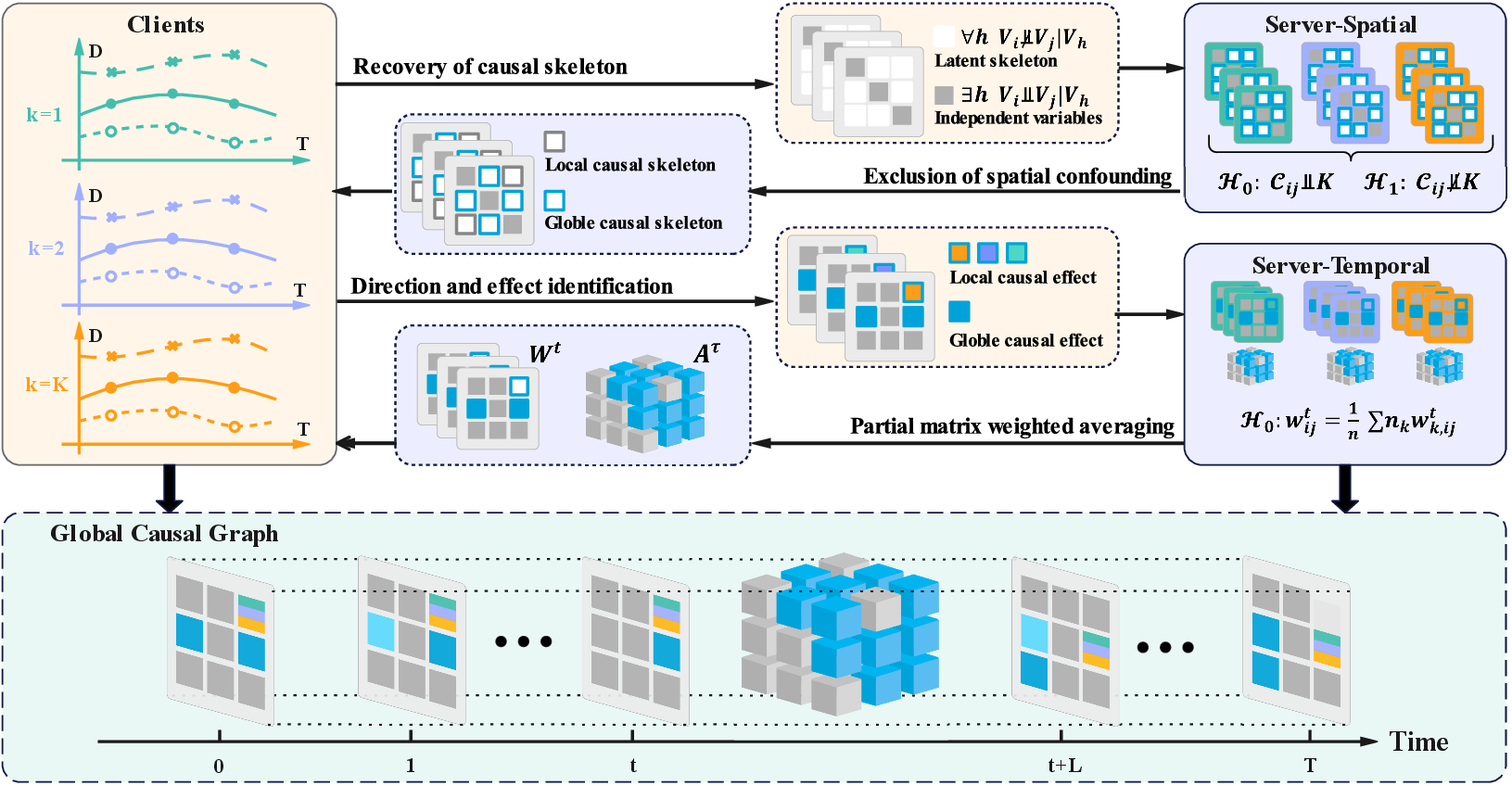}
\captionsetup{font={small}}
\caption{The framework of DisDy-ICPT}
\label{system}
\end{figure*} 


\section{Introduction}

\IEEEPARstart{C}{ausal} discovery plays a foundational role in many scientific and engineering domains: it exposes mechanisms that improve interpretability, supports counterfactual reasoning, and yields models that generalize more robustly under distribution shifts. Time-series data are particularly important in applications such as carbon monitoring, weather forecasting, transportation systems and biomedical monitoring, because their decisions depend on temporally structured cause–effect relations. Recovering causal structure from time series is however intrinsically harder than from i.i.d. data: one must handle both time-lagged (inter-slice) influences and instantaneous (intra-slice) effects, account for temporal dependence, and separate direct causal links from indirect or spurious associations induced by latent factors.

These difficulties are compounded in modern IoT and sensor networks where measurements are collected at many spatial locations. Data become decentralized across clients, and latent spatial factors (e.g., unobserved micro-climates, sensor calibration biases, local interventions) can induce client-specific confounding that varies across locations. In such settings a practical causal discovery system must satisfy three desiderata simultaneously: (i) model dynamic causal relationships across time (including changing contemporaneous structure), (ii) mitigate spatially varying unobserved confounders so that discovered relations are invariant and reliable, and (iii) respect data-locality and privacy constraints by operating in a federated (decentralized) fashion without sharing raw time series.

Prior work has progressed on these axes but not jointly. Constraint-based and invariant methods (e.g., CD-NOD \cite{cd-nod}) exploit distributional shifts to detect changing mechanisms, while recent time-series methods (e.g., DyCAST \cite{DyCAST}) model time-varying contemporaneous graphs by parameterizing causal evolution with Neural Ordinary Differential Equations (Neural ODEs). These approaches, however, assume centralized access to all data and are therefore ill-suited to decentralized, privacy-sensitive deployments. Conversely, federated causal learning methods such as FedCDH \cite{FCD} address distribution across clients but target static observational data and do not capture temporal dynamics and lagged causal effects. Moreover, existing federated approaches typically do not explicitly remove client-varying spatial confounders prior to causal structure search, which can lead to systematic false positives in learned graphs.

Motivated by these gaps, we propose Distributed Dynamic Invariant Causal Prediction in Time-series (DisDy-ICPT), a federated framework that jointly addresses temporal dynamics, spatial heterogeneity, and privacy. DisDy-ICPT operates in two distinct phases. First, a Phase I: Distributed Invariant Skeleton Mining (DISM) stage (Algorithm \ref{alg:DISM}) acts as a pre-processing step. It federates time-sliced kernel statistics to generate a set of robust, time-varying priors. For example, to rule out spurious correlations caused by a burst of noise at a certain client at a certain time affecting both variables, we design the statistic $\Omega$. To ensure efficiency, this is performed at a sparse sampling rate $T_S$. This stage outputs dynamic hard and soft constraints ($\mathbf{S}^{(t)}$, $\mathbf{L}_{\text{Soft}}^{\text{Final}, (t)}$) for the contemporaneous graph, and static constraints ($\mathbf{S}_{A}$, $\mathbf{L}_{\text{Soft}, A}^{\text{Final}}$) for the lagged graph. Second, a Phase II: Dynamic Causal Trajectory Optimization (DCTO) stage (Algorithm \ref{alg:DCTO}) uses a federated Neural ODE to learn the causal weights. This stage is deeply integrated with the priors from DISM, enforcing the hard constraints ($\odot \mathbf{S}$) structurally and using the soft constraints ($\odot \mathbf{L}_{\text{Soft}}$) as an adaptive $L_1$ penalty.

Theoretically, we show two complementary guarantees. Using characteristic kernels and concentration bounds for RKHS cross-covariance operators, we prove the detectability of client-varying confounding: under mild estimation error bounds and a separation condition on client operators, our DISM phase reliably flags heterogeneous dependence patterns that single-client tests would miss. Second, under standard smoothness and bounded-variance assumptions, we prove a FedAvg-style convergence bound for federated Neural ODE training: the convergence is guaranteed up to terms quantifying stochastic variance, heterogeneity drift, and solver bias.

We evaluate DisDy-ICPT on three classes of experiments. On controlled synthetic SEMs we design specific test cases to show that our DISM phase correctly identifies spatial confounding (for $\mathbf{S}^{(t)}$), spatial inconsistency (for $\mathbf{L}_{\text{Spatial}}^{(t)}$), and temporal instability via median filtering. On the realistic CausalTime benchmark we demonstrate superior AUROC/AUPRC for edge detection when environments are partitioned into clients. Finally, on real energy time-series we use the discovered causal structure to inform federated forecasting models and observe consistent improvements in MAE and RMSE versus federated black-box baselines. Ablation studies validate (i) the necessity of each constraint component (e.g., w/o $\mathbf{S}$, w/o $\mathbf{L}_{\text{Spatial}}$, (ii) the robustness of the Neural ODE causal parameterization, and (iii) the efficiency gains from our temporal sampling $T_S$.

To summarize, our main contributions are:
\begin{itemize}
\item We introduce DisDy-ICPT, the first federated framework that jointly learns dynamic causal structure in time series and mitigates client-varying spatial confounding while preserving data locality.
\item We propose a DISM procedure that generates dynamic ($\mathbf{S}^{(t)}$, $\mathbf{L}_{\text{Soft}}^{(t)}$) and static ($\mathbf{S}_{A}$, $\mathbf{L}_{\text{Soft}, A}$) causal priors based on federated KCI tests, a novel temporal smoothing logic, and efficient time-based sampling.
\item We develop a DCTO phase that integrates these dynamic and static priors into a latent Neural ODE with a unified parameter set $\theta$, which is then trained efficiently using Federated Averaging.
\item We perform extensive experiments on synthetic, simulated (CausalTime), and real datasets, including comprehensive ablation studies, demonstrating that DisDy-ICPT improves structural recovery and downstream predictive performance compared to state-of-the-art baselines.
\end{itemize}

\section{Related Work}

\begin{table*}[htbp]
\caption{Summary Table of Related Approaches}
\centering
\renewcommand{\arraystretch}{1.2}
\begin{tabularx}{\textwidth}{|>{\centering\arraybackslash}X|c|c|c|}
\hline
\textbf{Method} & \textbf{Distributed System} & \textbf{Time Dynamics} & \textbf{Client Heterogeneity} \\
\hline
CD‑NOD \cite{cd-nod} & $\times$ Centralized & $\checkmark$ Temporal shifts & $\checkmark$ Environmental shifts \\
\hline
SpaceTime \cite{SpaceTime} & $\times$ Centralized & $\checkmark$ Temporal shifts & $\checkmark$ Environmental shifts \\
\hline
CDANs \cite{ferdous2023cdans} & $\times$ Centralized & $\checkmark$ Dynamic lag/contemporaneous & $\times$ \\
\hline
DyCAST \cite{DyCAST} & $\times$ Centralized & $\checkmark$ Dynamic contemporaneous DAGs & $\times$ \\
\hline
FedCDH \cite{FCD} & $\checkmark$ Federated & $\times$ Static structure & $\checkmark$ Non-i.i.d clients \\
\hline
FedCDI \cite{abyaneh2022federated} & $\checkmark$ Federated (interventional) & $\times$ & $\checkmark$ Client-specific mechanisms \\
\hline
FedCausal \cite{yang2024federated} & $\checkmark$ Federated & $\times$ Static DAGs & $\checkmark$ \\
\hline
\textbf{DisDy-ICPT (Ours)} & \textbf{$\checkmark$ Federated} & \textbf{$\checkmark$ Dynamic causal structure} & \textbf{$\checkmark$ Spatial confounding} \\
\hline
\end{tabularx}
\label{RelatedWork}
\end{table*}
Prior research on dynamic causal discovery from time-series data has explored methods ranging from classical Granger causal and vector auto-regressive models to modern deep‑learning frameworks\cite{runge2019detecting,granger1969investigating}. Traditional techniques, such as VAR modeling and conditional independence tests like PCMCI, excel in mildly nonlinear and non-stationary settings but often assume homogeneous and centralized access to data\cite{ferdous2023cdans}. Recent neural approaches \cite{lowe2022amortized,sultan2022granger,shermin2009using} leverage shared dynamics across datasets to infer causal structure efficiently in time series, showing improved performance under noise and latent confounding. The current state-of-the-art method in this area is DYCAST\cite{DyCAST}, which models the evolving causal graph by integrating neural ODEs with continuous-time causal mechanisms. DYCAST captures fine-grained structural changes across time and achieves strong performance in identifying time-varying causal dependencies without requiring known environment labels or interventions. However, these methods, including DYCAST, primarily focus on temporal variations in causal effects within a single data source. They do not account for spatial causal heterogeneity arising from multiple distributed data sources or clients, where causal mechanisms may differ due to unobserved environmental or population factors. Addressing this  heterogeneity is essential for robust and generalizable causal discovery in federated or distributed settings.

Complementing dynamic models, a body of work on invariant causal prediction across environments has emerged from principles of Invariant Risk Minimization (IRM)\cite{arjovsky2019invariant} and Differentiable Invariant Causal Discovery (DICD)\cite{wang2022differentiable}. These frameworks leverage multi‑environment data to enforce structural stability in causal relations while discounting spurious associations that vary with environment\cite{peters2016causal}. However, few methods jointly address dynamics, environment invariant structure, and confounding control in spatiotemporal or environmental settings\cite{absar2021discovering}. For instance, state‐dependent approaches like State‑Dependent Causal Inference (SDCI) model nonstationarity by conditioning on latent states, offering identifiability guarantees for variable causal structures over time\cite{rodas2021causal}. Time‑series deconfounder approaches based on RNNs and latent variable modeling address time‑varying confounding in sequential treatments, but focus primarily on healthcare applications and assume rich proxy availability\cite{trifunov2022time}. More general environmental overviews show the need for causal methods that control confounding in spatiotemporal observational data, such as ambient pollution studies using spatial adjustment methods\cite{reich2021review,shimizu2006linear}. Spatial causal inference frameworks, especially in environmental and epidemiological domains, address confounding by exploiting spatial structure or incorporating spatial covariates into causal models, but typically without temporal dynamics or federated learning considerations.\cite{reich2021review,pfister2019invariant}

In distributed or federated learning, research in federated optimization has matured under central tasks like FedAvg and FedProx, designed to address client heterogeneity and communication efficiency\cite{kairouz2021advances,mcmahan2017communication,li2020federated}. There has also been exploratory work on distributed causal inference for static settings, such as protocols for federated causal effect estimation without sharing raw data\cite{meurisse2023federated,abyaneh2022federated}. Yet, these approaches generally assume i.i.d. samples across clients and static causal graphs, limiting applicability to temporally evolving environments\cite{li2025federated,yang2024federated}. In contrast, the most widely recognized federated causal discovery algorithm to date is FedCDH (Federated Causal Discovery from Heterogeneous Data), which extends the constraint-based CD-NOD framework to the federated setting. To address data heterogeneity across clients, FedCDH introduces a surrogate variable corresponding to the client index and develops a federated conditional independence test (FCIT) for causal skeleton discovery. It further proposes a federated independent change principle (FICP) to determine causal directions. CD-NOD itself supports nonstationary scenarios and is capable of modeling dynamic causal structures by identifying distributional changes over time. However, FedCDH fails to leverage this temporal modeling capability in its federated adaptation. In practice, it treats each time point or temporal segment as an independent data sample, overlooking the smooth, continuous transitions characteristic of time-series data. 

A smaller but growing literature has attempted to integrate temporal dynamics, causality invariance, and environmental context, particularly for earth‑system applications\cite{reichstein2019deep,su2024deep,alyousifi2020modeling}. Methods that combine Bayesian structural time‑series models (e.g. CausalImpact) allow for counterfactual inference in intervention contexts but remain centralized and static in design\cite{brodersen2015inferring}. Reviews of spatiotemporal causality in Earth science emphasize detecting nonlinear relationships and controlling environmental confounding, but rarely account for federated constraints\cite{frndak2023reducing}.

\begin{algorithm}
\caption{Algorithm 1: Distributed Invariant Skeleton Mining (DISM)}
\label{alg:DISM}
\begin{algorithmic}[1]
\linespread{1}\selectfont
\STATE \textbf{Input:} Distributed data $\{\mathbf{V}_{k}\}_{k=1}^K$, $\mathbf{Z}$, $\delta_{\text{hard}}$, sampling rate $T_S \ge 1$.
\STATE \textbf{Output:} Dynamic Priors $\{\mathbf{S}^{(t)}\}$, $\{\mathbf{L}_{\text{Soft}}^{\text{Final}, (t)}\}$; Static Priors $\mathbf{S}_{A}$, $\mathbf{L}_{\text{Soft}, A}^{\text{Final}}$.
\STATE \textbf{Client-side :}
\STATE $\mathcal{T}_{\text{sampled}} \leftarrow \{1, 1+T_S, 1+2T_S, \dots, T\}$
\FOR {client $k=1$ to $K$}
    \FOR {time $t \in \mathcal{T}_{\text{sampled}}$}
        \STATE Compute $\mathbf{Stat}_k^{(t)}$: $n_k^{(t)}$, $\mathcal{C}_{\mathcal{T}, k}^{(t)}$, $\mathbf{A}_{k}^{(\tau, t)}$.
    \ENDFOR
    \STATE \textbf{Upload} $\{\mathbf{Stat}_k^{(t)}\}_{t \in \mathcal{T}_{\text{sampled}}}$ to Server.
\ENDFOR

\STATE \textbf{Server-side:}
\STATE Initialize $\mathbf{S}^{(t)}$, $\mathbf{L}_{\text{Soft}}^{\text{Final}, (t)}$ for $t=1..T$
\STATE \textbf{Compute Dynamic Priors (for W):}
\STATE Initialize raw indicators $\mathbf{I}_{ij,k}^{(t)} \leftarrow 0$ for all $(i,j,k,t)$
\FOR {time $t \in \mathcal{T}_{\text{sampled}}$}
    \STATE $N^{(t)} \leftarrow \sum_k n_k^{(t)}$
    \STATE $\mathcal{C}_{\mathcal{T}}^{(t)} \leftarrow \sum_{k=1}^K \frac{n_k^{(t)}}{N^{(t)}} \mathcal{C}_{\mathcal{T}, k}^{(t)}$
    \STATE $\mathbf{S}^{(t)} \leftarrow \text{ComputeHardMask}(\mathcal{C}_{\mathcal{T}}^{(t)}, \mathbf{Z}, U, \delta_{\text{hard}})$
    \FOR {$k=1$ to $K$}
        \STATE $\mathbf{I}_{ij,k}^{(t)} \leftarrow \text{KCI\_Test}(\mathcal{C}_{\mathcal{T}, k}^{(t)}, V_i, V_j \mid \mathbf{Z})$
    \ENDFOR
\ENDFOR

\STATE \textbf{Correct, Interpolate, and Compute Priors:}
\STATE $\{\mathbf{I}_{\text{Corrected}, k}^{(t)}\} \leftarrow \text{TemporalFilter}(\{\mathbf{I}_{k}^{(t)}\}_{t \in \mathcal{T}_{\text{sampled}}})$
\FOR {$t=1$ to $T$}
    \STATE $t_{\text{prev}} \leftarrow \max(t' \in \mathcal{T}_{\text{sampled}} \mid t' \le t)$
    \STATE $\mathbf{S}^{(t)} \leftarrow \mathbf{S}^{(t_{\text{prev}})}$
    \STATE $\mathbf{L}_{\text{Soft}}^{\text{Final}, (t)} \leftarrow \text{ComputeSoftMask}(\mathbf{S}^{(t)}, \{\mathbf{I}_{\text{Corrected}, k}^{(t_{\text{prev}})}\})$
\ENDFOR

\STATE \textbf{Compute Static Priors (for A):}
\STATE $\mathbf{A}^{(\tau)} \leftarrow \sum_{t \in \mathcal{T}_{\text{sampled}}} \sum_{k=1}^K \frac{n_k^{(t)}}{N} \mathbf{A}_{k}^{(\tau, t)}$
\STATE $\mathbf{S}_{A} \leftarrow \text{ComputeHardMask}(\mathbf{A}^{(\tau)}, \mathbf{Z}, U, \delta_{\text{hard}})$
\STATE $\{\mathbf{I}_{k,A}\} \leftarrow \text{KCI\_Test}(\mathbf{A}_{k}^{(\tau, \cdot)}, \dots \mid \mathbf{Z})$
\STATE $\mathbf{L}_{\text{Soft}, A}^{\text{Final}} \leftarrow \text{ComputeSoftMask}(\mathbf{S}_{A}, \{\mathbf{I}_{k,A}\})$
\STATE \textbf{Return} $\{\mathbf{S}^{(t)}\}$, $\{\mathbf{L}_{\text{Soft}}^{\text{Final}, (t)}\}$, $\mathbf{S}_{A}$, $\mathbf{L}_{\text{Soft}, A}^{\text{Final}}$.
\end{algorithmic}
\end{algorithm}

\begin{algorithm}
\caption{Algorithm 2: Dynamic Causal Trajectory Optimization (DCTO)}
\label{alg:DCTO}
\begin{algorithmic}[1]
\linespread{1}\selectfont
\STATE \textbf{Input:} Dynamic Priors $\{\mathbf{S}^{(t)}\}, \{\mathbf{L}_{\text{Soft}}^{\text{Final}, (t)}\}$; Static Priors $\mathbf{S}_{A}, \mathbf{L}_{\text{Soft}, A}^{\text{Final}}$; Iterations $R$; Local steps $E$; Learning rate $\eta$.
\STATE \textbf{Output:} Optimized Dynamic $\{\mathbf{W}^{(t)}\}_{t=1}^T$ and Static $\{\mathbf{A}^{(\tau)}\}_{\tau=1}^{L}$.
\STATE \textbf{Server Initialization:}
\STATE $\theta^{0} \leftarrow \text{Initialize}(\dots)$ \COMMENT{Contains $\theta_{\text{dyn}}$ and $\mathbf{A}$}
\FOR {training iteration $r=0$ to $R-1$}
    \STATE \textbf{Broadcast} $\theta^r$ to all clients.
    \STATE \underline{Client $k$:}
    \STATE $\theta_{k} \leftarrow \theta^r$
    \FOR {local epoch $e=1$ to $E$}
        \STATE $\mathbf{A}_{\text{eff}, k} \leftarrow \theta_{k}[\mathbf{A}] \odot \mathbf{S}_{A}$
        \STATE $h^{(t_0)} \leftarrow \text{Encoder}(\mathbf{V}_k ; \theta_{k}[\phi_{\theta}])$
        \STATE $\frac{dh^{(t)}}{dt} = f_{\text{Base}}(h^{(t)}, \mathbf{W}_{\text{eff}}^{(t)}, \mathbf{A}_{\text{eff}, k}; \theta_{k}[\xi_{\theta}])$
        \STATE Solve $h^{(t)} \leftarrow \text{ODESolver}(\dots)$
        \STATE $\mathbf{W}_{\text{raw}}^{(t)} \leftarrow \text{Decoder}(h^{(t)} ; \theta_{k}[\psi_{\theta}])$.
        \STATE $\mathbf{W}_{\text{eff}}^{(t)} \leftarrow \mathbf{W}_{\text{raw}}^{(t)} \odot \mathbf{S}^{(t)}$
        \STATE $L_{\text{Soft\_W}} = \lambda_{W} \cdot \frac{1}{T} \sum_{t=1}^T \|\mathbf{W}_{\text{eff}}^{(t)} \odot \mathbf{L}_{\text{Soft}}^{\text{Final}, (t)}\|_1$
        \STATE $L_{\text{Soft\_A}} = \lambda_{A} \cdot \|\mathbf{A}_{\text{eff}, k} \odot \mathbf{L}_{\text{Soft}, A}^{\text{Final}}\|_1$
        \STATE $L_k = L_{\text{MSE}} + L_{\text{DAG}} + L_{\text{Soft\_W}} + L_{\text{Soft\_A}}$
        \STATE $\theta_{k} \leftarrow \theta_{k} - \eta \nabla_{\theta} L_k$
    \ENDFOR
    \STATE \textbf{Upload} updated parameters $\theta_{k}$.

    \STATE \underline{Server (FedAvg):}
    \STATE $\theta^{r+1} \leftarrow \sum_{k=1}^K \frac{n_k}{N} \theta_{k}$
\ENDFOR
\STATE $\mathbf{W}^{(t)} \leftarrow \text{Decode}(\theta^{R}[\psi_{\theta}], \dots)$
\STATE $\mathbf{A}^{R} \leftarrow \theta^{R}[\mathbf{A}]$
\STATE \textbf{Return} Optimized $\{\mathbf{W}^{(t)}\}_{t=1}^T$ and $\mathbf{A}^{R}$.
\end{algorithmic}
\end{algorithm}

\begin{algorithm}
\caption{Algorithm 3: DisDy-ICPT (Main Procedure)}
\label{alg:DisDy-ICPT_Main}
\begin{algorithmic}[1]
\linespread{1}\selectfont
\STATE \textbf{Input:} Distributed observational data $\{\mathbf{V}_{k}\}_{k=1}^K$.
\STATE \textbf{Output:} Optimized Dynamic Causal Matrices $\{\mathbf{W}^{(t)}\}_{t=1}^T$ and $\{\mathbf{A}^{(\tau)}\}_{\tau=1}^{L}$.
\STATE \textbf{Phase I: Distributed Invariant Skeleton Mining (DISM)}
\STATE $\{\mathbf{S}^{(t)}\}, \{\mathbf{L}_{\text{Soft}}^{\text{Final}, (t)}\}, \mathbf{S}_{A}, \mathbf{L}_{\text{Soft}, A}^{\text{Final}} \leftarrow \text{DISM}(\{\mathbf{V}_{k}\})$
\STATE \textbf{Phase II: Dynamic Causal Trajectory Optimization (DCTO)}
\STATE $\{\mathbf{W}^{(t)}\}, \{\mathbf{A}^{(\tau)}\} \leftarrow \text{DCTO}(\{\mathbf{S}^{(t)}\}, \{\mathbf{L}_{\text{Soft}}^{\text{Final}, (t)}\}, \mathbf{S}_{A}, \mathbf{L}_{\text{Soft}, A}^{\text{Final}})$
\STATE \textbf{Return} $\{\mathbf{W}^{(t)}\}, \{\mathbf{A}^{(\tau)}\}$.
\end{algorithmic}
\end{algorithm}

\section{Causal Discovery from Distributed Time Series}
\subsection{Problem Definition}

Consider $K$ clients, where the $k$‑th client holds $n_k$ samples and $n=\sum n_k $. Each sample $\mathbf{V}_{k,i}$ consists of $D$ time series variables of length $T$, collectively represented as $\mathbf{V}_{k,i} \in \mathbb{R}^{T \times D}$, where $\mathbf{V} = \{V_{1}, V_{2}, \dots, V_{D}\}$ and $V_{d} = \{V_{d}^1, V_{d}^2, \dots, V_{d}^T\}$ denotes the time series of the $d$-th variable. At each time step $t \in \{1, 2, \dots, T\}$, we define the system state as the vector $\mathbf{V}^t = [V_1^t, V_2^t, \dots, V_D^t] \in \mathbb{R}^D$, which captures the values of all $D$ variables at time $t$. According to the structural causal model (SCM) \cite{pearl2009causality}, each variable $V_d$ is generated by an additive noise model, which include a function $f_d$ of its parent variables $PA_d$ and some noise term $\epsilon_d$:

\begin{equation}
V_d = f_d(PA_d)+\epsilon_d.
\end{equation}

Extending this to the temporal setting, we establish an $L$-order time-lagged SCM for multivariate time series, where each variable at time $t$, $V_d^t$, depends on the values of its parent variables in previous time steps $t - \tau$, for $\tau = 1, \dots, L$, capturing both temporal and cross-variable causal dependencies.

\begin{equation}
    V_{d}^t = \sum_{t=0}^{L}{f_{d}^{t-\tau}(PA_{d}^{t-\tau})+\epsilon_{d}^{t-\tau}}
\end{equation}

This means that the earliest causes of $V^t_d$ occur $L$ time steps earlier. The notation $PA^{t-\tau}_d$ denotes the parents of $V^t_d$ in time slice $t - \tau$, and $f^{t-\tau}_d$ specifies the causal function mapping $PA^{t-\tau}_d$ to $V^t_d$. As in most causal‑discovery research, we assume that the noise term $\varepsilon^{t-\tau}_d$ is independent across time slices, clients, and samples, but it is related to variables and delay time steps. 

Ideally, the causal mechanism $f_i$ is consistent across clients. However, data heterogeneity across clients may arise due to unobserved confounding. To model this, we assume a set of latent confounding variables $U_k$. It is important to emphasize that here “heterogeneity” refers not to differences in data distributions, but to differences in the underlying causal mechanisms. For example, suppose that in every client we observe a random variable $V_{1}$ and two dependent variables defined by

\begin{equation}
V^t_{2} = 0.5\,V^t_{1} + 0.2\,a + b, 
\quad
V^t_{3} = 0.2\,V^t_{1} + 0.4\,a + c. 
\label{exp}
\end{equation}

In Client $k=1$ the parameters take the values $a = 0.1$, $b = 0.2$, and $c = 0.5$, whereas in Client $k=2$ they are $a = -0.1$, $b = -0.2$, and $c = -0.5$. In this scenario, the terms $b$ and $c$ act as environment‑specific causes and do not impede the identification of causal relationships among the $V$ variables. In contrast, the shared coefficient $a$ simultaneously influences both $V_{2}$ and $V_{3}$ and thus serves as a confounder; because $a$ cannot be effectively removed, it degrades the accuracy of causal discovery.

The combined effect of all variables in $\{U_k\}$ on variable $d$ at time $t$ is denoted by $\mathbf{U_k^t(d)}$. Concretely, if we absorb all unobserved confounding into a linear term $U_k(d)$, then for our two observed variables $V_2$ we have

\begin{equation}
\mathbf{U_1^t(2)}= 0.2\,a + b =0.22, 
\quad
\mathbf{U_2^t(2)}= 0.2\,a + b =-0.22\notag
\end{equation}

Accordingly, the client‑specific SCM can be rewritten as:
\begin{equation}
    V_{k,d}^t = \sum_{t=0}^{L}f_{d}^{t-\tau}(PA_{d}^{t-\tau} , \mathbf{{U^t_k(d)}} ,\epsilon_{d}^{-\tau})
\end{equation}
Regarding the dynamic variation of time and the confounding of spatial variation, we make the following assumptions based on previous work.
\begin{Assumption}(Spatial Confounder Sufficiency): \label{spatial Confounder}
\\
There is no confounder in the dataset of one domain, but the changes of different causal modules across different domains can be dependent.
\begin{equation}
{U_k(d)} \perp\!\!\!\perp i \mid k, \tau
\end{equation}
\begin{equation}=
\exists j \ne j',\quad U_k(d_j) \not\!\perp\!\!\!\perp U_k(d_{j'})
\end{equation}
\end{Assumption}
This assumption implies a form of modular heterogeneity across clients, which is although the causal mechanisms vary due to latent variables $U_k$, the induced statistical discrepancies across clients are sufficiently structured. Such structured heterogeneity provides identifiable signals for detecting confounded edges based on cross-client mechanism inconsistencies. 
\begin{Assumption}(Temporal Causal Invariance): \label{Pseudo}
\\
The contemporaneous causal structure can equivalently be viewed as evolving over time, while the lagged causal influences are assumed to remain temporally invariant.
\begin{equation}
f_{d}^{t}(PA_{d}^{t} , \mathbf{{U_k(d)}} ,\epsilon_{d}^{t}) \not\!\perp\!\!\!\perp t 
\end{equation}
\begin{equation}
\sum_{t=1}^{L}f_{d}^{t-\tau}(PA_{d}^{t-\tau} , \mathbf{{U_k(d)}} ,\epsilon_{d}^{t-\tau}) \perp\!\!\!\perp t
\end{equation}
\end{Assumption}
This temporal smoothness is consistent with the continuous-time parameterization in Neural ODEs, where the instantaneous causal mechanism $W^{t}$ evolves according to differentiable dynamic rule $\frac{dW^t}{dt}=f_\theta(w^t,t)$. Hence the assumed temporal continuity directly justifies the use of ODE-based trajectory learning in next phase. 

In this work, we present DisDy-ICPT (Distributed Dynamic Invariant Causal Prediction in Time-series), our two-stage federated algorithm designed to learn time-varying causal structures that are invariant across heterogeneous environments. Given $K$ clients, our goal is to recover the sequence of instantaneous adjacency matrices $\{\mathbf{W}^{(t)}\}_{t=1}^T$ and the static lagged adjacency tensors $\{\mathbf{A}^{(\tau)}\}_{\tau=1}^L$ without sharing raw data. DisDy-ICPT achieves this by separating the problem into two distinct phases, as detailed in Algorithm \ref{alg:DisDy-ICPT_Main}: Phase I: Distributed Invariant Skeleton Mining (DISM) (Algorithm \ref{alg:DISM}) to extract robust causal priors, and Phase II: Dynamic Causal Trajectory Optimization (DCTO) (Algorithm \ref{alg:DCTO}) to learn the dynamics via Neural ODEs.

\subsection{Phase I: Distributed Invariant Skeleton Mining (DISM)}
To facilitate subsequent analysis, we simplify the causal effects among variables by assuming a linear structural form. Specifically, the causal relationships among variables $\mathbf{V}^t$ at time $t$ can be written as:
\begin{equation}
\mathbf{V}^t = \mathbf{V}^t \mathbf{W}^t + \sum_{\tau=1}^{L} \mathbf{V}^{t-\tau} \mathbf{A}_\tau + \mathbf{\epsilon}^t,
\label{eq:linear_scm}
\end{equation}
where $\mathbf{W}^t$ denotes the instantaneous causal adjacency matrix at time $t$, and each $\mathbf{A}_\tau$ captures the time-lagged causal influence. Our objective is to learn the time-varying weights $\{ \mathbf{W}^t \}_{t=1}^T$ (which are spatially invariant but temporally dynamic) and the static weights $\{ \mathbf{A}_\tau \}_{\tau=1}^L$ (which are both spatially and temporally invariant), as defined in Assumption \ref{Pseudo}.

The goal of DISM is to generate the full set of priors required by DCTO. This includes the dynamic hard prior $\mathbf{S}^{(t)}$ and soft prior $\mathbf{L}_{\text{Soft}}^{\text{Final}, (t)}$ (for $\mathbf{W}^t$), and the static hard prior $\mathbf{S}_{A}$ and soft prior $\mathbf{L}_{\text{Soft}, A}^{\text{Final}}$ (for $\mathbf{A}$).

This phase is built upon the principles of federated kernel statistics aggregation, inspired by FedCDH \cite{FCD}. To handle non-linear relationships without sharing raw data, clients map their data into a high-dimensional feature space using Random Fourier Features (RFFs) $\Phi_w(\cdot)$ \cite{FCD}. The feature map $\Phi_w(\cdot): \mathbb{R}^D \to \mathbb{R}^h$ is defined as:
\begin{equation}
\Phi_w(\mathbf{V}) = \sqrt{2/h}[\cos(\mathbf{w}_1^T \mathbf{V} + b_1), \dots, \cos(\mathbf{w}_h^T \mathbf{V} + b_h)]^T.
\end{equation}
Crucially, to ensure computational tractability while capturing dynamics, we introduce a sampling rate $T_S$. We assume the causal structure is slowly varying (per Assumption \ref{Pseudo}), allowing us to compute constraints only at sampled time steps $\mathcal{T}_{\text{sampled}} = \{1, 1+T_S, \dots, T\}$. Clients compute a time-sliced local kernel covariance tensor $\mathcal{C}_{\mathcal{T}, k}^{(t)} \in \mathbb{R}^{D \times D \times h \times h}$ only for $t \in \mathcal{T}_{\text{sampled}}$. These, along with time-sliced lag statistics $\mathbf{A}_{k}^{(\tau, t)}$ and temporal variance $\mathbf{\Omega}_{k}^{(t)}$, are uploaded to the server.

The server then performs all computations in a pre-processing step. For the dynamic priors (for $\mathbf{W}^t$), the server iterates only over the sampled time steps $t \in \mathcal{T}_{\text{sampled}}$. It first aggregates the global covariance tensor for that specific time:
\begin{equation}
\mathcal{C}_{\mathcal{T}}^{(t)} = \sum_{k=1}^K \frac{n_k^{(t)}}{N^{(t)}} \mathcal{C}_{\mathcal{T}, k}^{(t)}.
\end{equation}
This time-sliced aggregation is the core mechanism enabling our search for spatially invariant, yet temporally dynamic, constraints. The server uses the Kernel-based Conditional Independence (KCI) test framework \cite{FCD}, adopting the Federated Conditional Independence Test (FCIT) statistic:
\begin{equation}
\mathcal{T}_{CI}^{(t)} \triangleq n^{(t)}\|\mathcal{C}_{\hat{X}Y|Z}^{(t)}\|_F^2,
\end{equation}
which is computed from the entries of $\mathcal{C}_{\mathcal{T}}^{(t)}$. The dynamic hard prior $\mathbf{S}^{(t)}$ is then generated by performing this FCIT for each sampled $t$. If $\mathcal{T}_{CI}^{(t)} < \delta_{\text{hard}}$, we set $\mathbf{S}_{ij}^{(t)} = 0$, structurally removing spatially confounded connections.

We adopted the FCIT statistic as implemented in FedCDH\cite{FCD}, thereby directly leveraging an established federated implementation of kernal-based conditional independence testing. DISM thus inherits the practical validated in FCIT, subject to the same operational conditions. To address the fact that FCIT does not explicitly account for temporal dependencies, we further introduce a new soft temporal constraint into DISM.

The dynamic soft prior $\mathbf{L}_{\text{Soft}}^{\text{Final}, (t)}$ is designed as a binary (0/1) penalty mask based on a two-step process.
First, the server computes the raw local indicators $\mathbf{I}_{ij,k}^{(t)}$ by performing a local KCI test on each uploaded $\mathcal{C}_{\mathcal{T}, k}^{(t)}$ for $t \in \mathcal{T}_{\text{sampled}}$.
Second, to account for temporal anomalies (e.g., measurement noise causing 00100 or 11011 spikes), we apply a temporal consistency filter to the sparse indicator series $\{\mathbf{I}_{ij,k}^{(t)}\}_{t \in \mathcal{T}_{\text{sampled}}}$ for each $(i,j,k)$ pair, using $\mathbf{\Omega}_{k}^{(t)}$ to guide the filtering strength. This yields a smoothed indicator series, $\mathbf{I}_{\text{Corrected}, k}^{(t)}$.
Third, the server propagates these constraints to all time steps. For any non-sampled time $t$, its constraints are set to those of the most recent sampled step $t_{\text{prev}}$ (a zero-order hold).

Finally, the soft mask is generated for all $t$ based only on the spatial inconsistency of these corrected and propagated indicators. A connection is penalized if, and only if, it was not removed by the hard constraint ($\mathbf{S}_{ij}^{(t)} = 1$) but was found to be independent on at least one client after temporal smoothing. This logic is formally expressed as:
\begin{equation}
    \mathbf{L}_{\text{Soft}, ij}^{\text{Final}, (t)} = \mathbf{S}_{ij}^{(t)} \cdot \left(1 - \min_{k} \mathbf{I}_{\text{Corrected}, ij,k}^{(t)}\right).
\label{eq:soft_mask_logic}
\end{equation}
This ensures $\mathbf{L}_{\text{Soft}, ij}^{\text{Final}, (t)} = 1$ only if the connection is permitted by $\mathbf{S}$ but found to be spatially inconsistent.

Finally, the static priors for the lag matrix $\mathbf{A}$ are computed, reinforcing Assumption \ref{Pseudo}. The server aggregates $\mathbf{A}_{k}^{(\tau, t)}$ over all sampled clients and time steps to create a single static tensor $\mathbf{A}^{(\tau)}$. It then applies the same invariance and (spatial) inconsistency logic to this static, aggregated data to generate a static hard constraint $\mathbf{S}_{A}$ and a static soft constraint $\mathbf{L}_{\text{Soft}, A}^{\text{Final}}$.

\subsection{Phase II: Dynamic Causal Trajectory Optimization (DCTO)}
After DISM, all priors (dynamic $\{\mathbf{S}^{(t)}\}, \{\mathbf{L}_{\text{Soft}}^{\text{Final}, (t)}\}$ and static $\mathbf{S}_{A}, \mathbf{L}_{\text{Soft}, A}^{\text{Final}}$) are broadcast to all clients and remain fixed. The DCTO stage (Algorithm \ref{alg:DCTO}) learns the dynamic contemporaneous weights $\{\mathbf{W}^{(t)}\}$ and the static lag weights $\{\mathbf{A}^{(\tau)}\}$. This phase is built upon the DyCAST framework \cite{DyCAST}, which models causal dynamics using a latent Neural ODE.

We adopt the Encoder-Process-Decoder architecture from DyCAST to efficiently model the dynamics \cite{DyCAST}. In our framework, all learnable parameters are contained within a single parameter set $\theta$, which consists of two parts: the dynamic parameters $\theta_{\text{dyn}}$ (for the Encoder $\phi_{\theta}$, Processor $\xi_{\theta}$, and Decoder $\psi_{\theta}$) and the static parameters $\mathbf{A}$ (the lag matrix itself). The Encoder maps the initial state to $h^{(t_0)}$, the Processor evolves $h^{(t)}$ via $\frac{dh^{(t)}}{dt} = f_{\text{Base}}(h^{(t)}, \dots)$, and the Decoder reconstructs the dynamic matrix $\mathbf{W}_{\text{raw}}^{(t)} = \psi_{\theta}(h^{(t)})$.

Our main contribution in this phase is the deep integration of the full DISM prior set.
First, we enforce hard constraints on both dynamic and static matrices. The effective weights are computed as:
\begin{equation}
    \mathbf{W}_{\text{eff}}^{(t)} = \mathbf{W}_{\text{raw}}^{(t)} \odot \mathbf{S}^{(t)} \quad \text{and} \quad \mathbf{A}_{\text{eff}} = \theta[\mathbf{A}] \odot \mathbf{S}_{A}.
\end{equation}
This operation structurally ensures the learned trajectory stays within the invariant manifolds defined by DISM. Because the mask $S^{(t)}$ is fixed and non-learnable, the Handamard product preserves differentiability with respect to $W^{(t)}_{raw}$. Consequently, the ODE solver and backpropagation remain valid, and gradients are simply zeroed for the removed edges. 
Second, we perform soft constraint integration by replacing the standard $L_1$ penalty \cite{DyCAST} with our adaptive soft-constraint losses:
\begin{equation}
    L_{\text{Soft\_W}} = \lambda_{W} \cdot \frac{1}{T} \sum_{t=1}^T \|\mathbf{W}_{\text{eff}}^{(t)} \odot \mathbf{L}_{\text{Soft}}^{\text{Final}, (t)}\|_1,
\end{equation}
\begin{equation}
    L_{\text{Soft\_A}} = \lambda_{A} \cdot \|\mathbf{A}_{\text{eff}} \odot \mathbf{L}_{\text{Soft}, A}^{\text{Final}}\|_1.
\end{equation}
This loss guides the optimization to apply an $L_1$ penalty only to those connections identified in Phase I as being unreliable.

The full objective function $L_k = L_{\text{MSE}} + \lambda_{\text{DAG}} + L_{\text{Soft\_W}} + L_{\text{Soft\_A}}$ is optimized using Federated Averaging (FedAvg). In each round $r$, clients load the global model $\theta^r$. They then perform $E$ local steps of gradient descent to produce an updated local parameter set $\theta_{k}$. These updated parameters are uploaded to the server, which performs a weighted average $\theta^{r+1} \leftarrow \sum_{k=1}^K \frac{n_k}{N} \theta_{k}$ to create the new global model for the next round. This two-stage approach allows DisDy-ICPT to robustly learn dynamic causal structures that are invariant across space.

...to be continued

\bibliographystyle{IEEEtran}
\bibliography{main}

\end{document}